\documentclass[10pt,twocolumn,letterpaper]{article}

\usepackage[pagenumbers]{iccv} 

\usepackage{float}
\usepackage{etoolbox}
\usepackage{placeins}
\usepackage{subcaption}
\usepackage{supertabular}
\usepackage{longtable}
\usepackage{booktabs}
\usepackage{lineno}
\usepackage[accsupp]{axessibility}

\definecolor{iccvblue}{rgb}{0.21,0.49,0.74}
\usepackage[pagebackref,breaklinks,colorlinks,allcolors=iccvblue]{hyperref}

\def\paperID{} 
\def\confName{ICCV}
\def\confYear{2025}

\title{Back Home: A Computer Vision Solution to Seashell Identification for Ecological Restoration}

\author{
Alexander Valverde\thanks{Project developed while the main author was working at FIFCO} \\
FIFCO\\
\and
Luis Solano\\
FIFCO\\
\and
André Montoya\\
FIFCO\\
}

\begin{document}
\maketitle
\begin{abstract}
Illegal souvenir collection strips an estimated five tonnes of seashells from Costa Rica’s beaches each year, yet once these specimens are seized their coastal origin—Pacific or Caribbean—cannot be verified easily because the lack of information, preventing their return when are confiscated by the local authorities. To solve this issue, we introduce BackHome19K, the first large-scale image dataset annotated with coast-level labels, and propose a lightweight pipeline that infers provenance in real time on a mobile-grade CPU. 
A similar PaDiM-inspired anomaly filter pre-screens uploads, increasing robustness to user-generated noise.  
On a held-out test set the classifier attains 86.3\,\% balanced accuracy, while the filter rejects 93\,\% of 180 out-of-domain objects with zero false negatives.  
Deployed as a web application, the system has already processed 70\,000 shells for wildlife officers in under three seconds per image, enabling confiscated specimens to be safely repatriated to their native ecosystems. The dataset is available at \href{https://huggingface.co/datasets/FIFCO/BackHome19K}{FIFCO/BackHome19K}.
\end{abstract}
    
\section{Introduction}

\begin{figure}
    \centering
    \includegraphics[width=\linewidth]{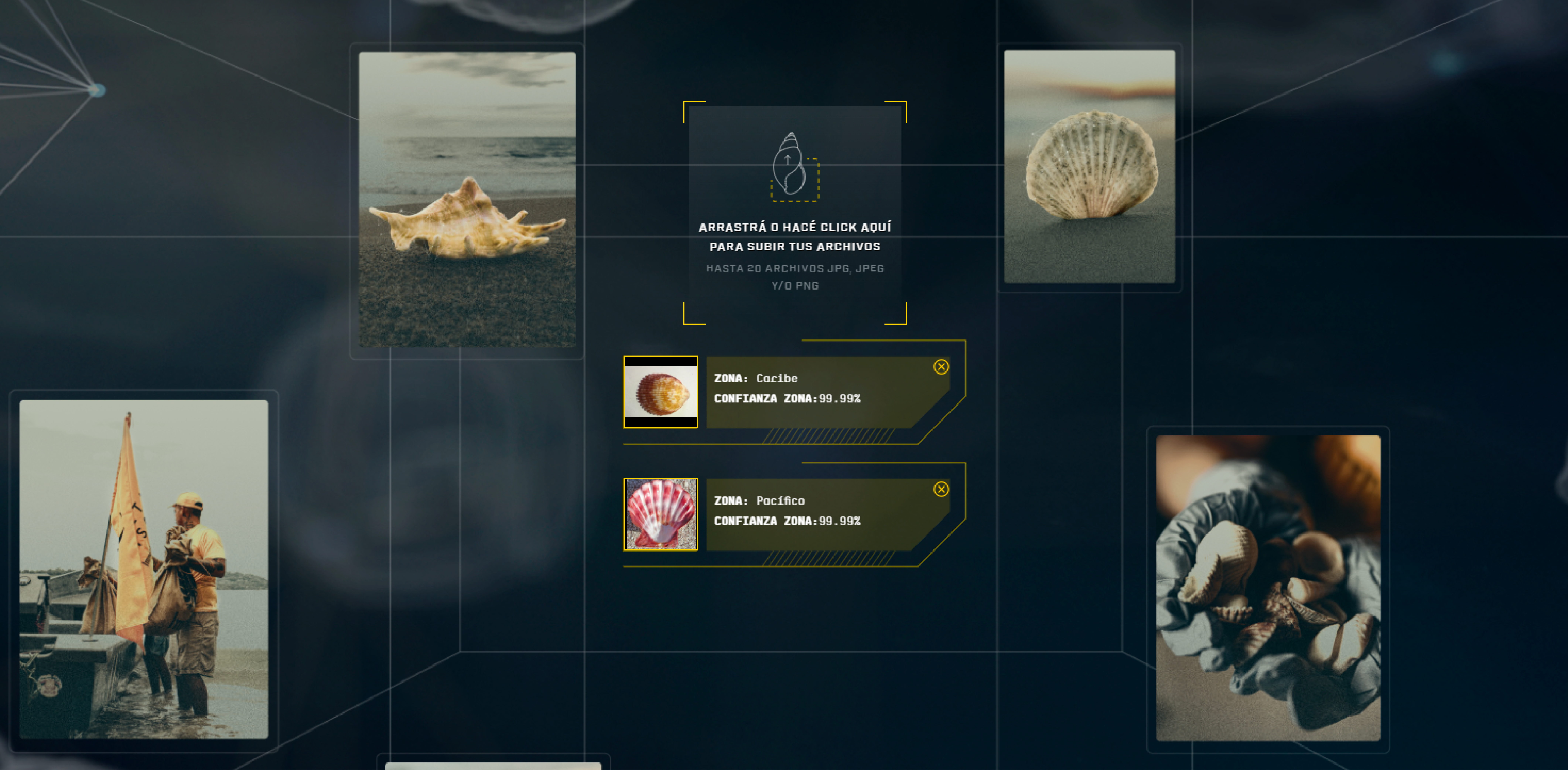}
    \caption{Web interface displaying the final output of our pipeline: for each uploaded seashell image the system returns its predicted coastal provenance with a confidence score}
    \label{fig:teaser}
\end{figure}
\vspace{-1pt}

\begin{figure*}[ht!]
    \centering
    \includegraphics[width=0.8\textwidth]{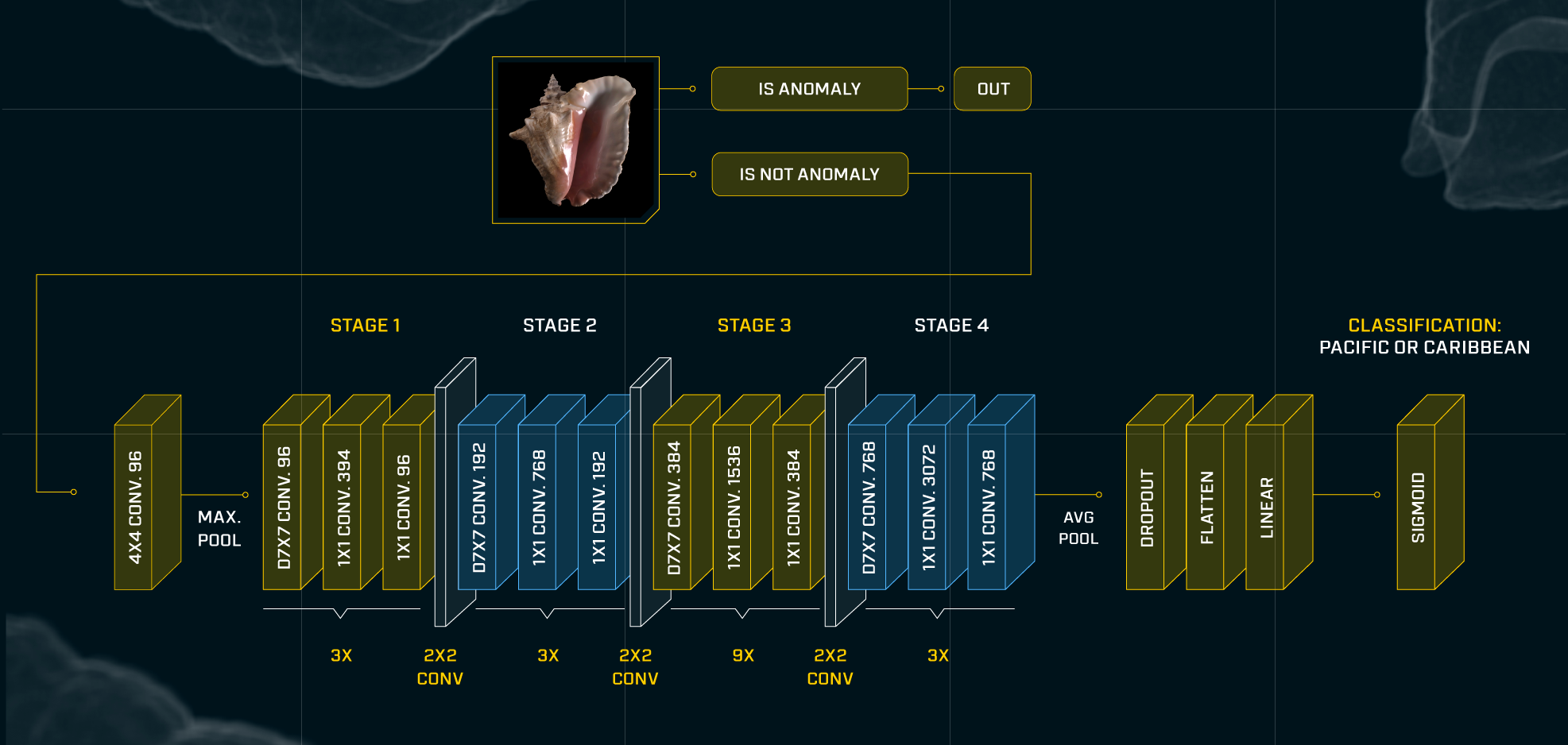}
    \caption{Overview of the proposed two-stage pipeline for seashell origin classification. The system first applies anomaly detection to identify and filter out abnormal specimens, then employs a classification network to predict the geographic origin of authentic seashells. The integrated approach ensures robust performance by eliminating outliers before classification.}
    \label{fig:pipeline}
\end{figure*}

Seashells are a keystone element of coastal ecosystems. Beyond providing habitat and shelter, their biological structures regulate sediment chemistry, buffer pH, and contribute to nutrient cycling. Recent materials-science studies \cite{cheng2023marine_shell, Kocot2016Seashell} have further shown that molluscan shells possess exceptional hardness, fracture toughness, corrosion resistance, and bio activity. Preserving this natural capital is therefore critical for biodiversity, shoreline stability, and sustainable innovation.

Despite their importance, seashell populations along Costa Rica’s beaches have sharply declined owing to unregulated souvenir collection by tourists. Customs officers at Juan~Santamaría International Airport routinely confiscate these shells, yet repatriation is impossible because the specimens’ provenance—Pacific or Caribbean coast—cannot be determined post-hoc. Returning them to the wrong habitat risks spreading invasive parasites, disrupting local gene pools, and skewing ecological surveys \cite{10.3389/fevo.2024.1454383, MARTINEZRUIZ2025109344}. Consequently, thousands of shells remain in storage each year instead of being restored to their native ecosystems \cite{costaricantimes2023leave, amcostarica2023four, ticotimes2023shells}

This conservation bottleneck presents a classic computer vision challenge: fine-grained classification where visually similar species from different ecosystems must be distinguished based on subtle morphological differences \cite{fishes9070267, article}. Traditional taxonomic identification focuses on species-level classification, but conservation applications require ecosystem-level provenance determination that demands specialized datasets and architectures optimized for geographic rather than purely biological distinctions.

The challenge extends beyond traditional fine-grained classification because Pacific and Caribbean shells often share nearly identical morphological features—similar color palettes, growth ring patterns, and overall silhouettes—yet originate from fundamentally different marine ecosystems \cite{gefaell2022shell, inproceedings}. This requires our model to learn extremely subtle distinguishing characteristics: minute texture variations, slight hue shifts, and microscopic geometric deviations that reflect different environmental conditions, water chemistry, and ecological pressures.

We tackle this logistical and conservation bottleneck through an image-based  classification model. Specifically, we formulate the task as a binary recognition problem—Pacific versus Caribbean—and introduce a large-scale dataset comprising $\sim$19\,000 high-resolution photographs containing 516 mollusk species presented in each location. Building on this resource, we design a lightweight convolutional neural network (CNN) tailored to the fine-grained visual cues that distinguish shells from the two coasts.

Our contributions are threefold:
\begin{enumerate}
    \item \textbf{Dataset.} We release \textbf{BackHome19K}, the first annotated image corpus of Costa Rican seashells with coast-level labels, captured in situ under natural lighting conditions across 516 species.
    \item \textbf{Model.} We propose a compact CNN architecture based on ConvNeXt-Tiny that balances classification accuracy with the low-latency constraints of field deployment.
    \item \textbf{System.} We integrate the model into a production-ready mobile and web application that incorporates anomaly detection for robust real-world performance, enabling wildlife authorities to triage confiscated shells in real time and streamline the path from seizure to ecosystem restoration.
\end{enumerate}

\section{Related Work}

Our research builds upon two primary areas: ecological classification models and image embeddings for real-time anomaly detection systems. The former involves techniques for identifying species or ecosystems based on visual or environmental data, while the latter focuses on methods for detecting outliers or non-conforming inputs to ensure reliability and robustness in real-world deployments.

\subsection{Architectures}

Convolutional Neural Networks (CNNs) have been central to image classification since the breakthrough by \cite{NIPS2012_c399862d}. Efficient variants like ResNet~\cite{he2016deep}, MobileNetV2~\cite{sandler2018mobilenetv2}, and DenseNet~\cite{huang2017densely} improved performance using residual connections, depthwise separable convolutions, and dense connectivity.

ConvNeXt~\cite{liu2022convnet} modernized CNNs by integrating transformer-inspired elements (large kernels, LayerNorm, GELU), narrowing the performance gap with transformers.

Vision Transformers (ViTs)~\cite{dosovitskiy2020image}, and later Swin~\cite{liu2021swin} and DeiT~\cite{touvron2021training}, replaced convolutions with attention-based mechanisms, achieving state-of-the-art results by modeling global patch relationships. This shift from CNNs to transformer-based models continues to improve recognition tasks across domains.

\subsection{Ecology Classification Models}

Initial research framed marine shell recognition as a fine-grained classification task. \cite{xue2021deepsea} introduced the DDI dataset and proposed Shuffle-Xception, enhancing accuracy on cluttered seabed scenes~\cite{xue2021ddi}. \cite{zhang2019shell} compiled a large benchmark (7,894 species, 59,000 images), evaluating traditional descriptors with classical classifiers.

To address class imbalance and visual similarity, \cite{yue2023flnet} proposed FLNet, a CNN with filter pruning and a hybrid loss tailored to skewed distributions. More recently, \cite{10.1371/journal.pone.0322711} applied transformers for fish recognition, leveraging transfer learning for high accuracy.

Despite progress, most approaches emphasize taxonomy and overlook ecological context—such as geographic provenance—which is crucial for conservation and real-world deployment.

\subsection{Image Embeddings and Anomaly Detection}

Deep learning enables encoding images as high-dimensional embeddings that capture semantic content, supporting tasks like retrieval, classification, and anomaly detection. These embeddings cluster semantically similar inputs in feature space, offering more meaningful comparisons than raw pixels. Self-supervised methods have largely superseded supervised CNNs by learning from unlabeled data.

SimCLR~\cite{chen2020simple} leveraged contrastive learning to encode image semantics via augmented views. CLIP~\cite{radford2021learning} extended this to multimodal embeddings, enabling zero-shot transfer across vision-language tasks. DINO~\cite{caron2021emerging} showed that vision transformers can learn structural and semantic features without labels.

For scalable retrieval, embeddings are indexed using approximate nearest neighbor (ANN) algorithms like FAISS, enabling efficient search in large datasets~\cite{medicalimageretrieval}.

Embeddings also underpin anomaly detection—key to identifying out-of-distribution (OOD) inputs. Similar techniques are used in NLP to detect hallucinations in LLMs~\cite{islam2024hallucination, su2024real_time, su2024entity}. Visual anomaly detection methods fall into three categories: (i) reconstruction-based (autoencoders, VAEs~\cite{sakurada2014anomaly, an2015vae}), (ii) probabilistic modeling (PaDiM~\cite{defard2021padim}, CS-Flow~\cite{rudolph2022csflow}), and (iii) confidence-based methods like ODIN~\cite{liang2020enhancingreliabilityoutofdistributionimage}.

While not central to our system, anomaly detection modules improve robustness against unexpected user-generated inputs in field conditions.

\section{Method}

This work makes three key contributions. First, we curate and release the first large-scale seashell image corpus, encompassing photographs from seashell species collected from both Costa Rica’s Pacific and Caribbean coasts—an essential resource for future marine-biodiversity studies. Second, we trained a lightweight ConvNeXt-Tiny–based classifier that distinguishes between the two ecosystems even for visually similar species from the same family that inhabit different shores. Third, we wrap the model in a production-ready web service that delivers predictions in under three seconds per image and incorporates an integrated anomaly detector, ensuring corrupted or out-of-distribution inputs are flagged before they reach the classifier.

\subsection{Dataset Construction}

We introduce \textbf{BackHome19K}, the first comprehensive seashell dataset specifically designed for ecosystem-level classification. Our dataset addresses a critical gap in existing marine datasets~\cite{zhang2019shell,matthewacs_south_florida_sea_shells}, which focus on taxonomic classification rather than bio geographic origin inference.

\begin{figure}[ht!]
    \centering
    \includegraphics[width=\columnwidth]{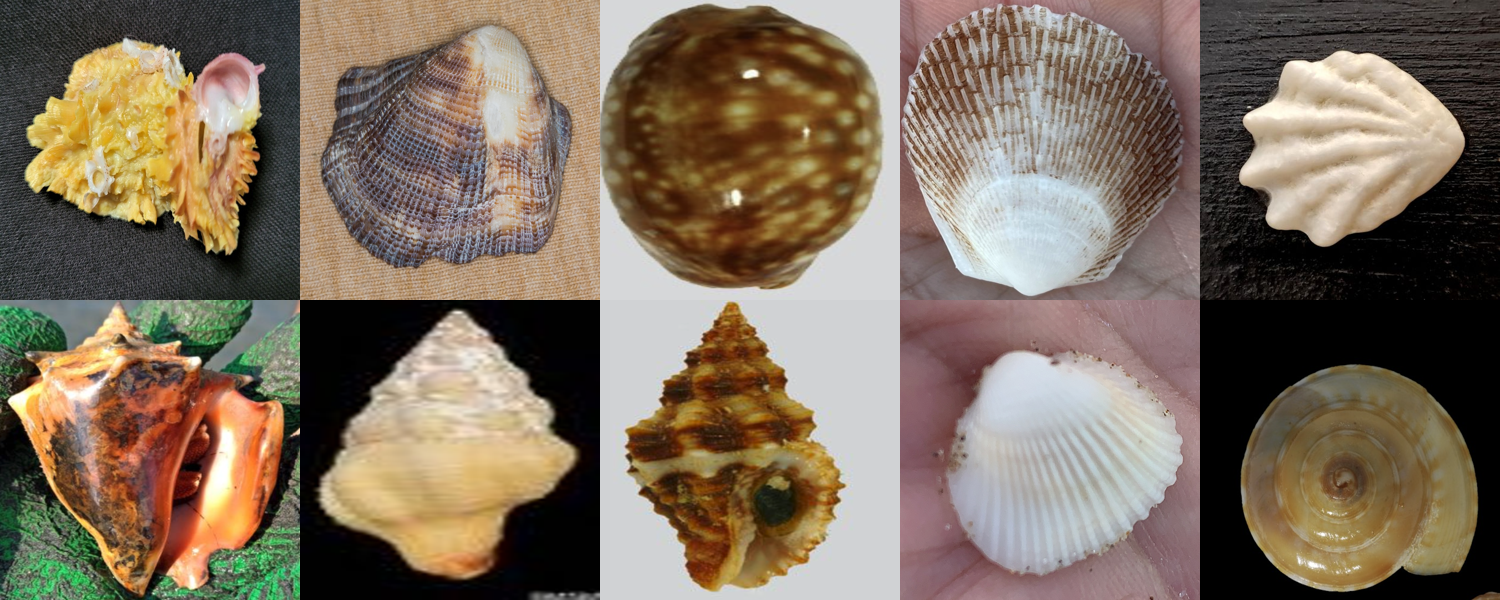}
    \caption{Representative examples of bivalves and gastropods collected from Caribbean coasts.}
    \label{fig:caribbean}
\end{figure}    

Because Pacific and Caribbean shells often share near-identical color palettes, growth rings, and overall silhouettes, a small dataset would leave the model unable to tease apart these subtle cues. We therefore assembled a broad and balanced collection spanning 516 species based on a list of seashells provided by the Biology School from the Universidad de Costa Rica.

Our dataset encompasses a diverse collection of marine mollusks from Pacific and Caribbean waters, representing major taxonomic groups across multiple families. Among the gastropods, we include specimens from families such as Acmaeidae (limpets), Buccinidae (whelks), Bullidae (bubble shells), Calliostomatidae (top shells), Cassidae (helmet shells), Cypraeidae (cowries), and Fasciolariidae (spindle shells), among numerous others. The bivalve collection features representatives from Arcidae (ark shells), Ostreidae (oysters), Pectinidae (scallops), Tellinidae (tellins), Semelidae (semele clams), and Veneridae (venus clams), providing comprehensive coverage of both filter-feeding and burrowing species across varied marine habitats.

\begin{figure}[ht!]
    \centering
    \includegraphics[width=\columnwidth]{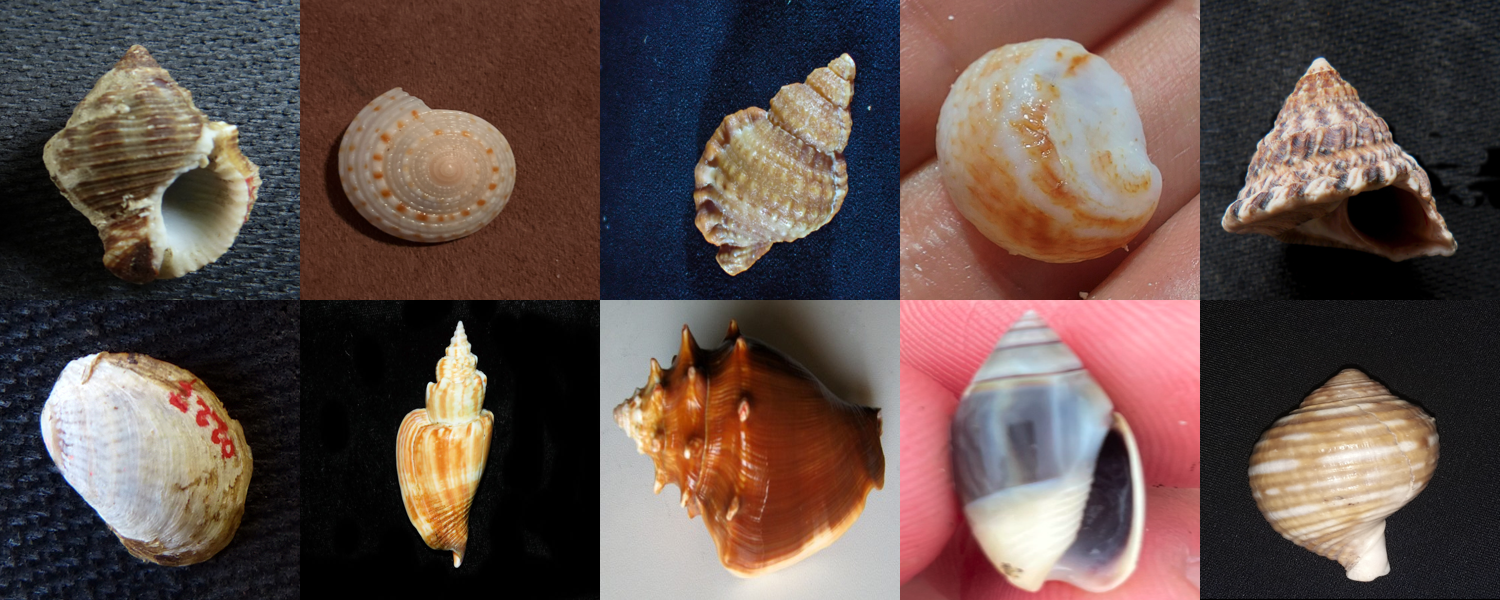}
    \caption{Representative examples of bivalves and gastropods collected from Pacific coasts.}
    \label{fig:pacific}
\end{figure}   

We systematically cataloged the species endemic to Pacific and Caribbean coasts through consultation with marine biology experts and cross-referencing multiple authoritative databases~\cite{conchology,inaturalist,fit_shells,conchyliNet}. Species selection criteria included: (1) confirmed presence in target ecosystems, (2) sufficient morphological distinctiveness for visual classification, and (3) availability of high-quality reference imagery. 

\begin{figure*}[ht!]
    \centering
    \includegraphics[width=0.8\textwidth]{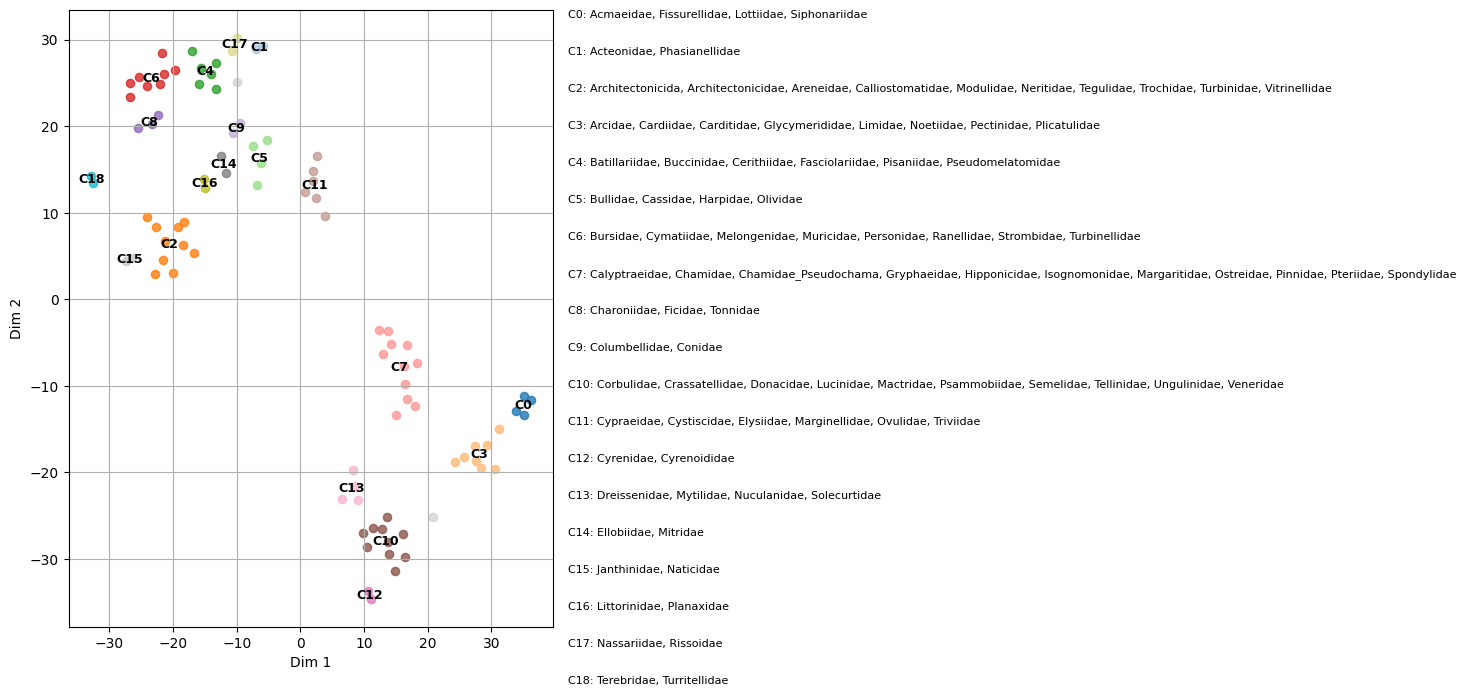}
    \caption{Seashell family clusters based on the mean feature vectors for each specie using embedding representations from all the images in the dataset using t-SNE \cite{vanDerMaaten2008tsne} and DBSCAN \cite{10.5555/3001460.3001507}}
    \label{fig:relations}
\end{figure*}    

The dataset was assembled by three researchers under the guidance of Dr. Yolanda Camacho García, who provided the species list. Using her taxonomic expertise, the lead researcher defined a labeling protocol to ensure consistency within classes and clear separation between them. Images were chosen based on key morphological traits—texture, color, shape, and distinctive external features—and organized into class‑specific folders grouped by ecosystem (Pacific vs. Caribbean). In Figure \ref{fig:relations}, we then computed each family’s mean embedding and applied t‑SNE to visualize clusters of families with similar color, shape, texture, and size.

\begin{figure*}[ht!]
    \centering
    \includegraphics[width=0.8\textwidth]{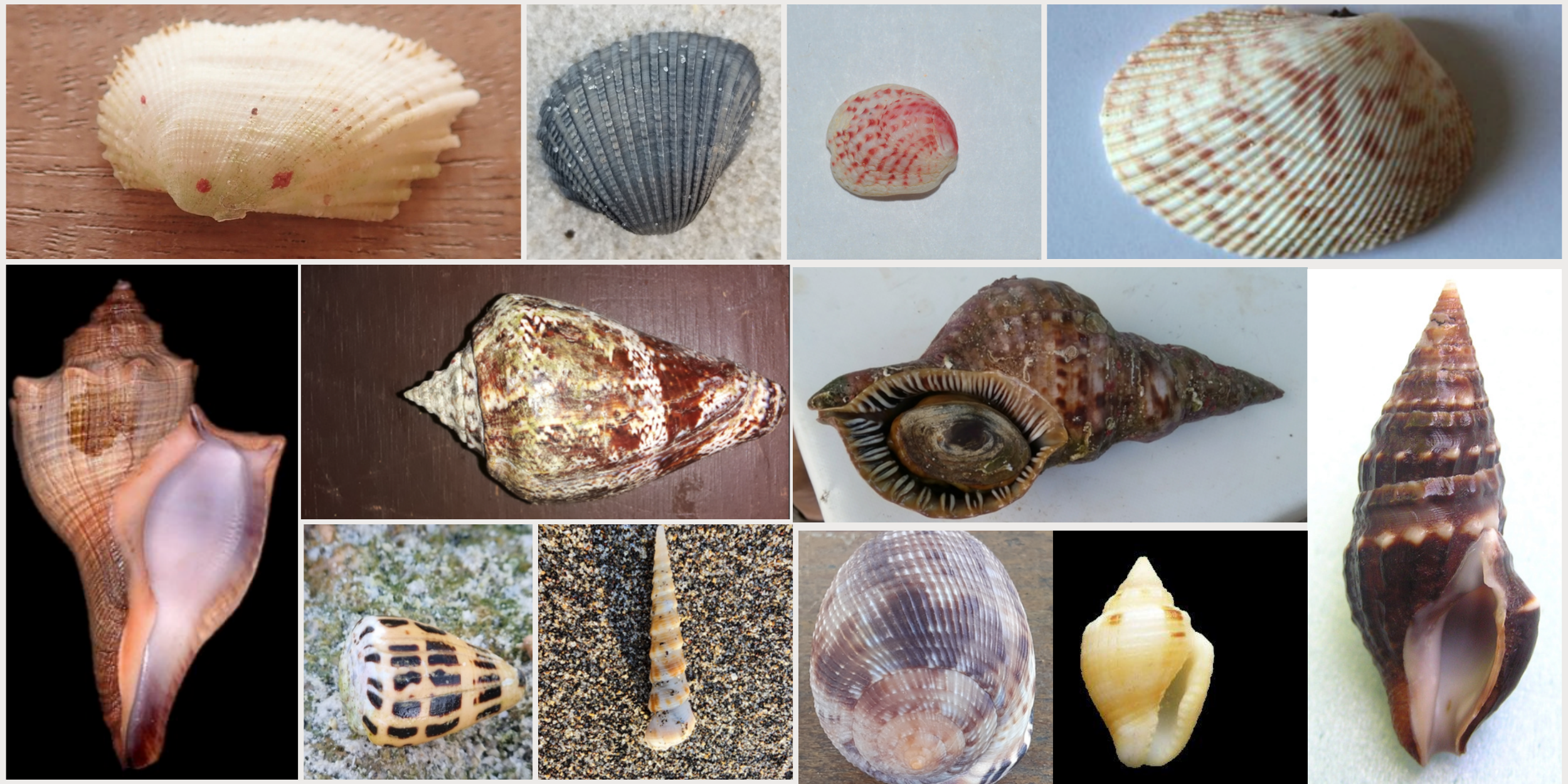}
    \caption{Representative specimens from our dataset showing morphological diversity across families. Top row (left to right): Pectinidae (scallop with characteristic radial ribs), Cardiidae (cockle with prominent radial sculpture), Cardiidae (heart cockle with distinctive coloration), Veneridae (venus clam with concentric growth lines). Middle row: Fasciolariidae (tulip shell with elongated spire), Cypraeidae (cowrie with glossy, oval shell), Turbinidae (turban shell with nacreous interior visible). Bottom row: Mitridae (mitre shell with geometric pattern), Turritellidae (tower shell with high spire), Tellinidae (tellin with smooth, elongated form), Buccinidae (whelk with pointed spire), Fasciolariidae (spindle shell with fusiform shape).}    \label{fig:seashells}
\end{figure*}    

Over 10 months, we assembled 19{,}051 images. Each of them underwent rigorous quality control including: taxonomic verification, removal of synthetic/composite images and standardization of lighting conditions To ensure morphological diversity, we collected 30–40 pictures per species showing different growth stages, orientations, and preservation states. 

The dataset was divided into Pacific and Caribbean subsets for the specific purpose of training the model introduced in the next section. However, the images are also organized by family, genus, and species, providing broader opportunities for future research. This structure enables studies focused on inter-species similarities and supports classification tasks based on taxonomic attributes rather than geographic origin.

\begin{table}[t]
\centering
\caption{Dataset composition and characteristics for Pacific and Caribbean divided into Gastropods and Bivalves}
\label{tab:dataset_stats}
\begin{tabular}{lcc}
\toprule
\textbf{Metric} & \textbf{Pacific} & \textbf{Caribbean} \\
\midrule
Total Species & 237 & 279 \\
Total Images & 9,505 & 9,553 \\
Gastropod Species & 130 & 149 \\
Bivalve Species & 107 & 130 \\
Avg. Images/Species & 40.1 & 34.2 \\
\bottomrule
\end{tabular}
\end{table}

\subsection{Model Architecture}

We adopt ConvNext-Tiny~\cite{liu2022convnet} as our classification backbone, initialized with ImageNet-1K pre-trained weights. Our architectural choice addresses two critical deployment constraints: computational efficiency for real-time inference and model size limitations for mobile deployment scenarios.

ConvNext modernizes traditional CNN design by incorporating key innovations from Vision Transformers while maintaining computational efficiency. The architecture employs large 7$\times$7 convolution kernels in early stages, enabling effective global feature capture without the quadratic complexity of self-attention mechanisms. This design proves particularly advantageous for fine-grained classification tasks where subtle morphological differences must be detected across varying spatial scales. We deliberately exclude Vision Transformers from consideration due to model size, ViT-Base requires 86M parameters \vs\ ConvNext-Tiny's 28M, creating storage constraints for field deployment; second, self-attention scales quadratically with input resolution, limiting real-time performance on resource-constrained devices.

This architecture already encodes a rich hierarchy of visual cues, from global shape and size in its early layers to increasingly fine color bands and micro-textures in its final block. By freezing the stem and first three stages we lock in the generic edge, contour, and chromatic primitives that distinguish broad shell geometries, while selectively unfreezing only the last block to relearn the millimeter-scale ridges, growth rings, and pigment speckles that separate near-identical families from Costa Rica’s Pacific and Caribbean coasts. 

\subsection{Anomaly Detection}

Real-world deployment necessitates filtering non-seashell inputs that could degrade classification performance. Therefore, our approach focuses on using a vector representation of each training image and storing these vectors in a vector database. This enables efficient comparison against images provided by volunteers, resulting in a more accurate classification based on their similarities. This method ensures that only images containing the seashell as the primary element of interest are selected, reducing the impact of extraneous noise that could adversely affect the classification. Such an approach is particularly beneficial for a web application utilized by a wide range of users, many of whom may lack professional or scientific photography skills.

Our anomaly–detection module adopts the embedding logic of PaDiM \cite{defard2021padim} while using a more lightweight backbone.  Concretely, each image is forwarded through SqueezeNet~1.0 \cite{iandola2016squeezenetalexnetlevelaccuracy50x}; in line with PaDiM we take the global-average-pooled activations of the final convolutional layer as the feature vector, yielding a 1\,000-dimensional descriptor. 

The choice of SqueezeNet is driven by its compactness and representational strength. With only $\approx 1.2$\,M parameters—roughly $20 \times$ fewer than canonical backbones such as ResNet-50—it fits comfortably on resource-constrained devices. 
In addition, its Fire modules (1\,$\times$\,1 “squeeze” followed by 1\,$\times$\,1 / 3\,$\times$\,3 “expand” filters) and delayed down-sampling keep high-resolution feature maps deep into the network, enabling it to capture the fine local textures that reveal subtle, pixel-level anomalies.

\enspace Having obtained a 1\,000-dimensional embedding
$\mathbf e_q$ for each query image, the system then measures its affinity
to the dataset by computing the mean cosine similarity to its $k$
nearest neighbors:
\[
S \;=\; \frac{1}{k}\,\sum_{i=1}^{k}
      \frac{\langle \mathbf e_q , \mathbf e_i \rangle}
           {\lVert \mathbf e_q \rVert\,\lVert \mathbf e_i \rVert}.
\]

If \(S < \lambda\), where \(\lambda\) is a predefined threshold, the image is classified as an anomaly:

\[
\text{Classification} =
\begin{cases} 
\text{Valid Image} & \text{if } S \geq \lambda \\
\text{Anomaly} & \text{if } S < \lambda
\end{cases}
\]

This method ensures that only valid seashell images are processed by the classification system, improving overall accuracy and reliability. The threshold $\lambda$ was determined empirically by analyzing the distribution of similarity scores between known seashell images in our dataset.

\begin{figure}[ht!]
    \centering
    \includegraphics[width=\columnwidth]{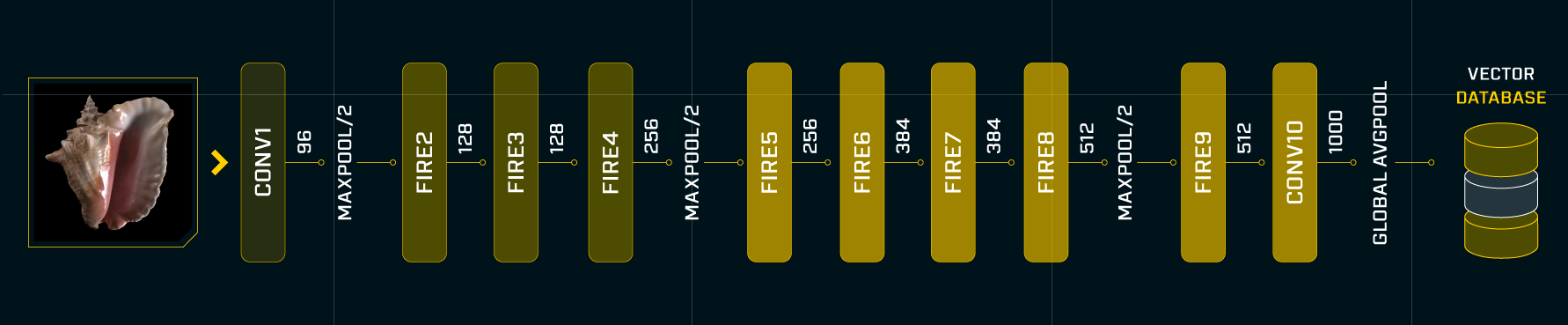}
    \caption{Feature extraction and embedding generation from input images using the SqueezeNet architecture}
    \label{fig:squeezenet}
\end{figure}

\section{Experiments}

In the following section we indicate the procedure for the model training and hyperparameter selection.

\subsection{Dataset Setting}

The dataset was divided into three subsets for training, validation, and testing, with 70\% allocated to the training set for model training, 15\% to the validation set for hyperparameter tuning and performance evaluation during training, and 15\% to the test set for evaluating the final model performance on unseen data. This split ensured a balanced representation of families across all subsets, supporting a rigorous and reliable evaluation of the model’s performance. The decision to retain almost 2,900 pictures for testing was made because of the need to ensure that the model would perform correctly on completely new seashells, as it was required to be used in a real-time framework with completely new seashells, ensuring that our dataset distribution could cover those undiscovered seashells. In addition, all images were resized to 224x224 pixels to maintain consistent input dimensions with the model architecture.

To improve the diversity of the training data and help the model perform better in real-world conditions, we applied several data augmentation techniques during training. These included: (1) geometric transformations, such as random rotations (up to $\pm45^\circ$), horizontal and vertical flips, and zooming in or out (scaling between 0.8$\times$ and 1.2$\times$) to reflect different angles at which seashells might be photographed; (2) photometric changes, including variations in brightness (up to $\pm20\%$), contrast (up to $\pm15\%$), and saturation (up to $\pm10\%$), to simulate differences in lighting; (3) spatial adjustments, such as random cropping (while keeping the aspect ratio) and slight elastic distortions, mimicking natural variations in how seashells may appear. These augmentations made the training set more realistic and better suited for field deployment, especially in cases where the images are taken by non-experts under uncontrolled conditions.

\subsection{Implementation Details and Hyperparameter Analysis}

All experiments were implemented in PyTorch and conducted on a single NVIDIA A100 GPU. We performed extensive hyperparameter optimization across seven distinct configurations, systematically evaluating the impact of unfrozen layer counts, training duration, and learning rate scheduling on model performance.

We evaluated three optimizers (AdamW \cite{loshchilov2019decoupled}, Adam \cite{kingma2015adam}, and Stochastic Gradient Descent (SGD) \cite{robbins1951stochastic}) with initial learning rates spanning $1\times10^{-4}$ to $1\times10^{-2}$, comparing two scheduling strategies: cosine annealing and ReduceLROnPlateau. Our final configuration employs SGD with an initial learning rate of 0.001 and cosine annealing, which demonstrated superior convergence characteristics.

The architectural decisions were guided by empirical findings:
\begin{itemize}
\item Unfreezing the final 30 layers of ConvNext-Tiny provided optimal feature adaptation without overfitting
\item Strategically placed dropout layers mitigated overfitting on underrepresented seashell families
\item The chosen configuration particularly benefits fine-grained classification, where subtle inter-class differences demand careful feature extraction
\end{itemize}

This systematic exploration establishes a robust baseline for seashell classification while providing insights into transfer learning optimization for fine-grained visual tasks.

\section{Results}

\subsection{Classification Performance}
As shown in table \ref{tab:performance_metrics}, the model achieves balanced performance across both ecosystems, with slightly higher accuracy for Caribbean specimens (87.10\% \vs\ 85.43\%). This difference may reflect the larger number of Caribbean species in our training set (279 \vs\ 237 Pacific species) .

\begin{table}[t]
\centering
\caption{Top-1 test accuracy (mean ± std across ten experiments) of four backbone CNNs. ConvNeXt-Tiny outperforms the next-best DenseNet121 by roughly 6 percentage points.}
\label{tab:baseline_comparison}
\begin{tabular}{lc}
\toprule
\textbf{Architecture} & \textbf{Test Accuracy (\%)} \\
\midrule
ResNet50 & 78.3 $\pm$ 0.4 \\
DenseNet121 & 80.2 $\pm$ 0.3 \\
MobileNetV2 & 79.3 $\pm$ 0.5 \\
ConvNext-Tiny & \textbf{86.28} $\pm$ 0.5 \\
\bottomrule
\end{tabular}
\end{table}

ConvNeXt-Tiny's superior performance stems from its modern architectural design, which effectively preserves high-frequency visual details crucial for fine-grained classification tasks. The network's large 7×7 convolution kernels in early stages enable comprehensive feature capture, while its hierarchical structure maintains discriminative micro-textures and subtle morphological variations that distinguish shells from different ecosystems.

In contrast, traditional architectures like ResNet50 \cite{he2016deep} and MobileNetV2 \cite{sandler2018mobilenetv2} struggle to preserve these fine-grained details due to their aggressive down-sampling strategies and smaller receptive fields. This limitation becomes particularly problematic when classifying morphologically similar seashells from the same taxonomic family but different geographic origins, where the distinguishing features may be limited to minute surface textures.

\begin{table}[!t]
  \centering
\caption{Classification performance on BackHome19K test set (2,858 images) showing balanced accuracy across Pacific and Caribbean ecosystems}  
\label{tab:performance_metrics}
  \setlength{\tabcolsep}{4pt}        
  \scriptsize                        
  \begin{tabular*}{\columnwidth}{@{\extracolsep{\fill}}lcccc}
    \toprule
    Ecosystem & Accuracy (\%) & Precision (\%) & Recall (\%) & F1 (\%) \\
    \midrule
    Pacific     & 85.43 & 88.70 & 85.43 & 87.00 \\
    Caribbean   & 87.10 & 83.45 & 87.10 & 85.29 \\
    \midrule
    \textbf{Overall} & \textbf{86.28} & \textbf{86.23} &
    \textbf{86.13} & \textbf{86.17} \\
    \bottomrule
  \end{tabular*}
\end{table}

The results demonstrate that the model performs consistently across both ecosystems, achieving a balanced accuracy of 85.43\% on Pacific samples and 87.10\% on Caribbean samples. This indicates that the system correctly classifies nearly 9 out of 10 seashells into their respective native ecosystems, highlighting its robustness and generalization capabilities across diverse marine environments.

\subsection{Anomaly Detection Performance}

Our anomaly detection system successfully filters non-seashell inputs while preserving legitimate specimens. Testing on 200 images across 20 object categories (10 images each, except 40 seashell images), the system achieved 100\% recall on seashell images while correctly identifying 90.5\% of non-seashell objects as anomalies using the empirically determined threshold $\lambda = 0.955$ and $k = 5$

To probe the robustness of our similarity–based filter, we curated an out-of-domain corpus from different standard vision benchmarks such as COCO \cite{lin2015microsoftcococommonobjects}, ImageNet-1k \cite{deng2009imagenet}, and Places365 \cite{zhou2017places}.
For each of 18 everyday object categories, we sampled ten high-quality photos, yielding 180 non-seashell test images.  
We also added forty previously unseen in-domain seashell pictures for a recall check.  

\begin{figure}[ht!]
    \centering
    \includegraphics[width=\columnwidth]{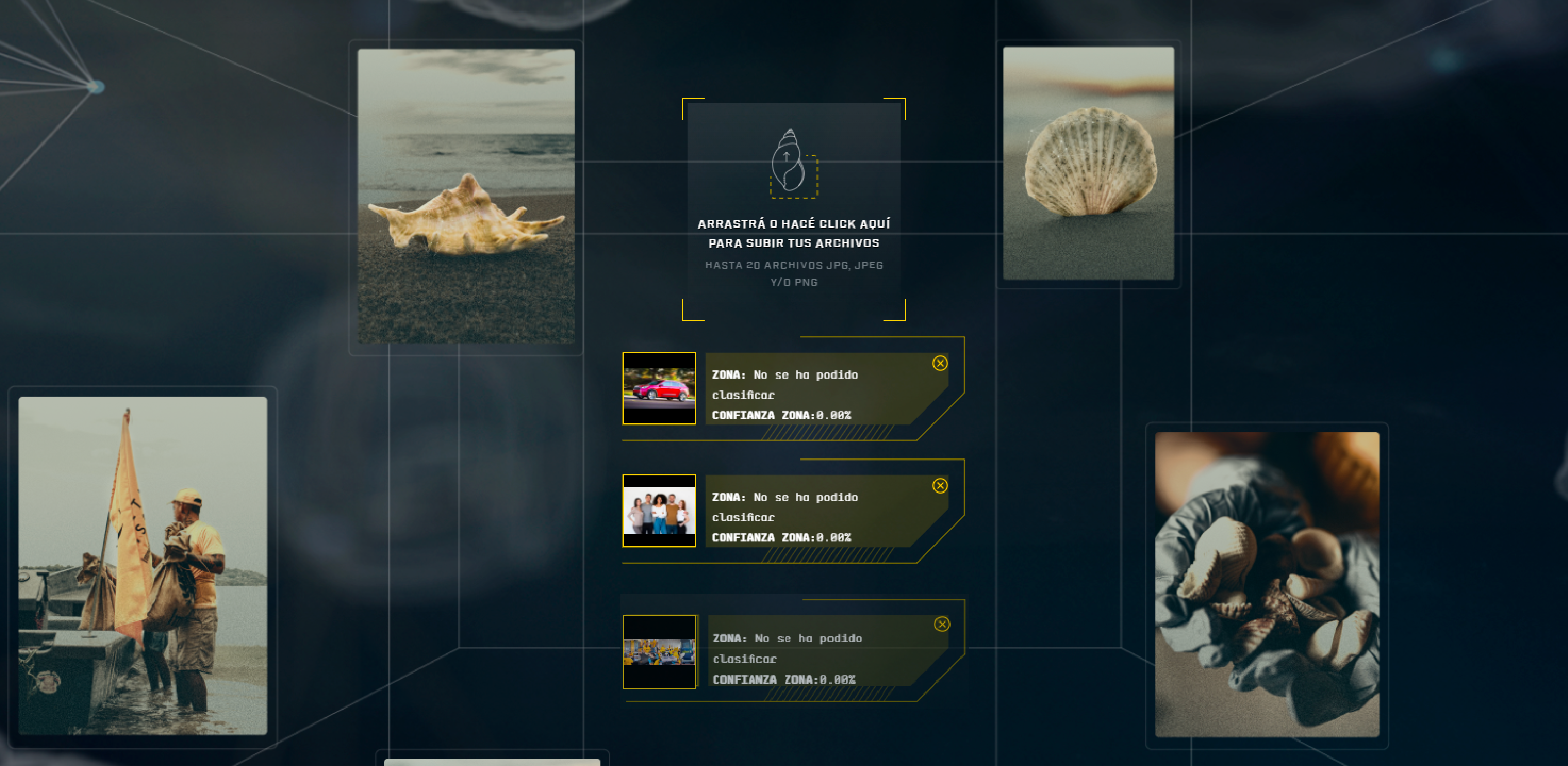}
    \caption{Anomaly detection interface displaying rejection of non-seashell objects using similarity-based filtering with threshold $\lambda = 0.955$, achieving 93\% precision on out-of-domain images}
    \label{fig:anomaly}
\end{figure}

A fixed threshold of 0.955 rejected 168 of the 180 off-domain images (93 \% true-positive rate).  
Perfect rejection was achieved for visually distinctive classes such as cats, cars, dogs, trucks, and backgrounds (10 / 10 each).  
More heterogeneous textures—notably reptiles, people, and frogs—accounted for the remaining misses, yet still exceeded a 60 \% rejection rate.  
Crucially, all forty seashell controls scored above the threshold, producing zero false negatives and confirming that the detector preserves recall on in-domain data while aggressively filtering other objects.

\section{Web Application}

To make the classification model truly accessible, we released a production‐ready two tier web service that lets users drag-and-drop shell photos and receive provenance predictions in \(\le\!3\) s. On the client side, a React application served via FastAPI "render" service, which exposes a receive-files endpoint: incoming images undergo Base64 encoding before being bundled with a JWT and asynchronously proxied, via HTTP/2 and Uvicorn's event loop, to a second FastAPI sevice on a predict endpoint. That service forwards the payload to a custom prediction endpoint on Google Cloud Platform. The predict service then streams the JSON response received from Vertex AI back to the frontend, which dynamically updates the UI.

The entire stack, React assets, FastAPI render and predict containers, is Dockerized and deployed to Cloud Run, leveraging scale to zero, sub second cold starts, and horizontal auto scaling. This architecture ensures sub 3 seconds end to end response times under heavy concurrency while minimizing idle costs.

\begin{figure}[ht!]
    \centering
    \includegraphics[width=\columnwidth]{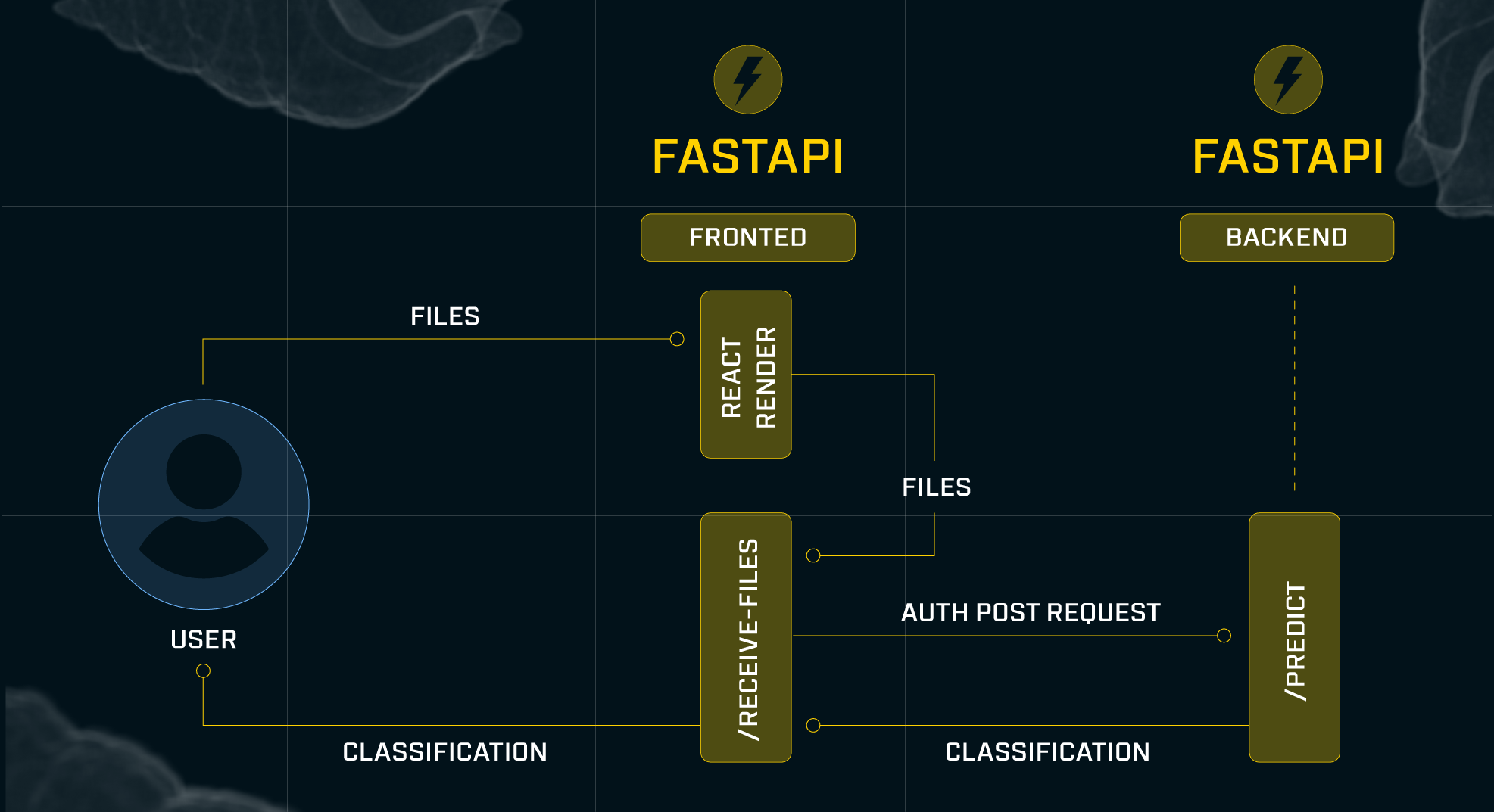}
    \caption{End-to-end technical pipeline for real-time shell classification system with sub-3-second response times via distributed FastAPI services and Cloud Run deployment}
    \label{fig:tech_pipeline_results}
\end{figure}

Cold-start latency on Cloud Run is \(\approx\!1.2\) s, while warm inference adds a median \(1.6\) s, yielding a 95th-percentile end-to-end time of \(<\!3\) s from upload to prediction display.

During a three-day public launch, the app served \(\sim\!200\) unique users who collectively classified \(\approx\!36\,000\) seashells averaging \(6.7\) images s\(^{-1}\) at peak traffic—without a single unhandled error.

\section{Ablations}

To evaluate the performance of the ConvNext-Tiny architecture and our decission to unfreeze a set of layers given the efficiency and considering the benefits of initial ones, we ran a set of experiments using the ConvNext-Tiny architecture introduced above with a set of same hyperparameters to evaluate the changes. We trained these models with SGD \cite{robbins1951stochastic} and with a learning rate of $1 \times 10^{-2}$ and just differing the learning rate epoch scheduler and the number of epochs. 

\begin{table}[htbp]
\centering
\caption{Ablation Study: Model Performance with Different Configurations to observe unfrozen layers importance}
\label{tab:ablation_study}
\small
\begin{tabular}{|c|c|c|c|}
\hline
\textbf{Unfrozen} & \textbf{Epochs} & \textbf{Scheduling} & \textbf{Accuracy} \\
\hline
0 & 50 & 25 & 83.24 \\
0 & 50 & 25 & 84.11 \\
10 & 100 & 50 & 80.15 \\
10 & 150 & 75 & 84.56 \\
10 & 250 & 125 & 84.16 \\
13 & 150 & 75 & 84.16 \\
\textbf{30} & \textbf{100} & \textbf{68} & \textbf{86.28} \\
\hline
\end{tabular}
\end{table}

Based on Table~\ref{tab:ablation_study}, we observe that unfreezing 30 layers yields the best performance with 86.28\% accuracy. Notably, this configuration required only 100 training epochs compared to the 13-layer unfrozen experiment, which trained for 150 epochs yet achieved 2.12\% lower accuracy (84.16\%). The accuracy is presented on the test dataset.

The results demonstrate a clear relationship between the number of unfrozen layers and model performance. Experiments with 10 unfrozen layers consistently underperformed compared to the 30-layer configuration, with accuracies ranging from 80.15\% to 84.56\%. While the baseline frozen models (0 unfrozen layers) achieved reasonable performance (83.24\%--84.11\%), they were still outperformed by the optimal unfrozen configuration.

It is important to note that the varying number of epochs across experiments was determined based on validation accuracy convergence patterns observed during training using an early stop function. Each configuration was trained until the validation accuracy plateaued, indicating that additional epochs would incur unnecessary computational costs without performance gains.
\section{Limitations}

While the classification system demonstrated strong performance, several limitations were observed. The model occasionally failed under suboptimal lighting conditions or unconventional viewing angles/distances, significantly affecting extracted visual features and reducing accuracy. Additionally, some Caribbean seashells lack sufficient training images in research databases, creating classification difficulties.

Although the anomaly detection mechanism filtered many irrelevant inputs, it was not infallible. Non-seashell objects with similar textures (coral fragments, rocks, marine debris) were sometimes misclassified as seashells, indicating the need for better fine-grained visual distinction.

\begin{figure}[ht!]
    \centering
    \includegraphics[width=\columnwidth]{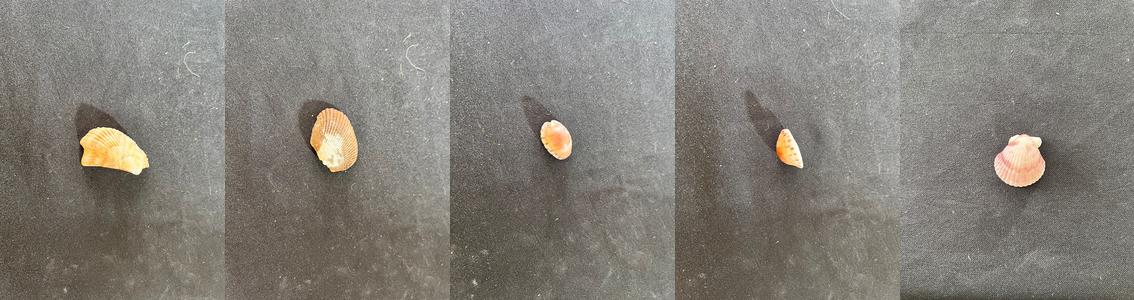}
    \caption{Example of seashells that were misclassified into Caribbean due to challenging conditions such as viewpoint, depth, lighting, and lack of visible texture or detail. From left to right: \textit{Pectinidae}, \textit{Arcidae}, \textit{Cypraea} (×2), and a small \textit{Cardiidae} shell.}

    \label{fig:tech_pipeline}
\end{figure}

The current dataset includes 516 species based on available scientific knowledge, though this initial version may not cover all existing species. Future updates will address these gaps as new discoveries emerge.

\section{Conclusion}
This research demonstrates that modern computer vision can effectively support conservation efforts through practical, deployable systems. Our BackHome19K dataset and classification pipeline achieve 86.3\% accuracy in determining seashell ecosystem provenance, enabling the automated return of confiscated specimens to their native habitats.

This work establishes a framework for ecosystem-level classification that could extend to other marine organisms and geographic regions. The integration of anomaly detection with fine-grained classification proves essential for real-world deployment, where user-generated content varies significantly in quality and relevance.

The public release of BackHome19K will enable comparative studies and methodological improvements across the computer vision for ecology community.

\section*{Acknowledgments}

We would like to thank FIFCO and the advertising agency Joystick for the funding and support provided throughout the duration of this project. We are especially grateful to Yolanda Camacho, PhD for her invaluable work and dedication in compiling the seashell lists. We also express our appreciation to the \textit{Sistema Nacional de Áreas de Conservación} (SINAC), the \textit{Universidad de Costa Rica} (UCR), and AERIS for their collaboration and assistance throughout this process. Finally, we extend our thanks to Vivian Mayorga from Joystick for her support in designing the web application and visual elements of this paper.

\onecolumn

\section*{Supplementary Material}

\subsection*{Anomaly Detection Algorithm}

\begin{table}[ht]
\begin{tabular}{lc}
\hline
\textbf{Category} & \textbf{Images Below Threshold} \\
\hline
Cats & 10/10 \\
People & 7/10 \\
Buildings & 8/10 \\
Cars & 10/10 \\
Trees & 10/10 \\
Rooms & 9/10 \\
Cows & 9/10 \\
Hospitals & 8/10 \\
Horses & 9/10 \\
Dogs & 10/10 \\
Backgrounds & 10/10 \\
Ships & 9/10 \\
Birds & 9/10 \\
Frogs & 7/10 \\
Trucks & 10/10 \\
Airplanes & 9/10 \\
Reptiles & 6/10 \\
Electronic Devices & 9/10 \\
Insects & 8/10 \\
Seashells & 0/40 \\
\hline
\end{tabular}
\caption{Anomaly Detection Performance Across Object Categories (n=10 images per category except seashells with n=40). * Threshold score = 0.955. Images with mean similarity scores below threshold are classified as anomalies.}\label{tab:anomalies}
\end{table}

\subsection*{List of Species}
\label{species:list_of_species}

The dataset was constructed using species from four distinct taxonomic and geographic groups: Caribbean Gastropods, Pacific Gastropods, Caribbean Bivalves, and Pacific Bivalves. The following lists provide the complete taxonomic nomenclature for each species included in the analysis, as provided by Yolanda Camacho García, PhD, Universidad de Costa Rica (UCR).

\begin{longtable}{p{\textwidth}}
\caption{Seashell Species - Caribbean Gastropoda} \label{tab:caribe_gastropoda} \\ 
\toprule
\textbf{Species} \\ 
\midrule
Acmaeidae Lottia antillarum, Acteonidae rictaxis punctostriatus, Architectonicidae Philippia krebsi, \\ 
Areneidae Arene cruentata, Buccinidae Hesperisternia karinae, Buccinidae Pollia auritula, \\ 
Buccinidae pisania auritula, Bullidae Bulla punctulata, Bullidae bulla mabillei, \\ 
Bullidae bulla striata, Bursidae Bursa cubaniana, Bursidae bursa thomae, \\ 
Calliostomatidae Calliostoma jujubinum, Calyptraeidae Bostrycapulus aculeatus, \\ 
Calyptraeidae Crepidula convexa, Calyptraeidae Ergaea walshi, Calyptraeidae crepidula aculeata, \\ 
Cassidae Cypraecassis testiculus, Cassidae Semicassis granulata, Cassidae cassis tuberosa, \\ 
Cerithiidae Cerithium eburneum, Cerithiidae cerithium guinaicum, Cerithiidae cerithium litteratum, \\ 
Cerithiidae cerithium lutosum, Charoniidae Charonia variegata, Columbellidae columbella mercatoria, \\ 
Columbellidae mazatlania fulgurata, Columbellidae mitrella ocellata, \\ 
Columbellidae nitidella laevigata, Columbellidae nitidella nitida, Columbellidae parametaria ovulata, \\ 
Conidae Conasprella mindana, Conidae Conus cardinalis, Conidae Conus mus, \\ 
Conidae Conus regius, Conidae Conus spurius lorenzianus, Conidae conus daucus, \\ 
Conidae conus jaspideus, Conidae conus spurius phlogopus, Cymatiidae Monoplex nicobaricus, \\ 
Cypraeidae Luria cinerea, Cypraeidae Macrocypraea zebra, Cypraeidae Naria acicularis, \\ 
Cypraeidae cypraea acicularis, Cypraeidae cypraea cinerea, Cypraeidae cypraea zebra, \\ 
Cystiscidae persicula interruptolineata, Ellobiidae Melampus coffea, Elysiidae elysia ornata, \\ 
Fasciolariidae Fasciolaria tulipa, Fasciolariidae Hemipolygona carinifera, \\ 
Fasciolariidae Poligona angulata, Fasciolariidae Polygona infundibulum, \\ 
Fasciolariidae latirus angulatus, Fasciolariidae leucozonia nassa, \\ 
Fasciolariidae leucozonia ocellata, Fissurellidae Diodora listeri, Fissurellidae Fissurella barbadensis, \\ 
Fissurellidae Hemitoma octoradiata, Fissurellidae diodora jaumei, \\ 
Fissurellidae fissurella angusta, Fissurellidae fissurella fascicularis, \\ 
Fissurellidae fissurella nodosa, Fissurellidae fissurella rosea, \\ 
Fissurellidae hemitona octoradiata, Fissurellidae lucapina aegis, \\ 
Fissurellidae lucapina sowerbii, Fissurellidae lucapina suffusa, Harpidae Morum oniscus, \\ 
Hipponicidae hipponix antiquatus, Littorinidae Cenchritis muricatus, Littorinidae Echinolittorina ziczac, \\ 
Littorinidae Littoraria angulifera, Littorinidae echinolittorina meleagris, \\ 
Littorinidae echinolittorina tuberculata, Littorinidae littoraria tessellata, Littorinidae littorina ziczac, \\ 
Littorinidae nodilittorina angustior, Marginellidae prunum holandae, Melongenidae melongena melongena, \\ 
Modulidae modulus modulus, Muricidae Stramonita floridiana, Muricidae Stramonita rustica, \\ 
Muricidae Vasula deltoidea, Muricidae chicoreus florifer, Muricidae chicoreus pomum, \\ 
Muricidae muricopsis deformis, Muricidae muricopsis oxytatus, Muricidae plicopurpura patula, \\ 
Muricidae thais haemastoma floridana, Muricidae thais rustica, Nassariidae Nassarius consensus, \\ 
Nassariidae nassarius albus, Nassariidae phos antillarum, Naticidae Natica marochiensis, \\ 
Naticidae Polinices hepaticus, Naticidae naticarius canrena, Naticidae polinices lacteus, \\ 
Naticidae sinum perspectivum, Neritidae Nerita fulgurans, Neritidae Nerita peloronta, \\ 
Neritidae nerita pelonronta, Neritidae nerita tessellata, Neritidae nerita versicolor, \\ 
Neritidae vitta virginea, Olividae Oliva fulgurator, Olividae oliva reticularis, \\ 
Olividae olivella minuta, Olividae olivella nivea, Ovulidae Cyphoma gibbosum, \\ 
Ovulidae Cyphoma signatum, Phasianellidae eulithidium bellum, Phasianellidae tricolia tessellata, \\ 
Pisaniidae Engina turbinella, Pisaniidae pisania pusio, Planaxidae Supplanaxis nucleus, \\ 
Planaxidae planaxis nucleus, Pseudomelatomidae Crassispira harfordiana, Ranellidae Cabestana labiosa, \\ 
Ranellidae Charonia tritonis, Ranellidae Monoplex nicobaricus, Ranellidae Monoplex pileare, \\ 
Ranellidae Monoplex vespaceus, Ranellidae Septa occidentalis, Rissoidae rissoina elegantissima, \\ 
Rissoidae rissoina sagraiana, Strombidae Aliger gigas, Strombidae Lobatus raninus, \\ 
Strombidae strombus pugilis, Tegulidae Agathistoma viridulum, Tegulidae Cittarium pica, \\ 
Tegulidae tegula excavata, Terebridae hastula cinerea, Tonnidae tonna galea, \\ 
Tonnidae tonna maculosa, Triviidae trivia nix, Triviidae trivia pediculus, Trochidae Calliostoma javanicum, \\ 
Turbinellidae Turbinella angulata, Turbinellidae Volutella muricata, Turbinidae Astraea caelata, \\ 
Turbinidae Lithopoma tectum, Turbinidae astraea phoebia, Turbinidae astraea tecta, \\ 
Turbinidae turbo cailletii, Turbinidae turbo castanea, Vitrinellidae solariorbis corylus, \\ 
Vitrinellidae vitrinella elegans, Volutidae voluta virescens \\ 
\bottomrule
\end{longtable}

\vspace{1cm}

\begin{longtable}{p{\textwidth}}
\caption{Seashell Species - Pacific Gastropoda} \label{tab:pacific_gastropoda} \\ 
\toprule
\textbf{Species} \\ 
\midrule
Architectonicida Heliacus areola bicanaliculatus, Architectonicidae Architectonica karsteni, \\ 
Architectonicidae Heliacus areola bicanaliculatus, Architectonicidae Heliacus caelatus, \\ 
Batillariidae Rhinocoryne humboldti, Bullidae Bulla punctulata, \\ 
Bursidae Alanbeuella corrugata, Bursidae Bursa rugosa, Bursidae Dulcerana granularis, \\ 
Bursidae bufonaria rana, Bursidae bursa granularis, \\ 
Calyptraeidae Bostrycapulus aculeatus, Calyptraeidae Calyptraea chinensis, \\ 
Calyptraeidae Calyptraea conica, Calyptraeidae Crepidula lessonii, \\ 
Calyptraeidae Crepidula marginalis, Calyptraeidae Crepidula striolata, \\ 
Cassidae Cypraecassis coarctata, Cerithiidae Cerithium adustum, \\ 
Cerithiidae Cerithium atromarginatum, Cerithiidae Cerithium browni, \\ 
Cerithiidae Cerithium muscarum, Columbellidae Anachis boivini, \\ 
Columbellidae Anachis lyrata, Columbellidae Anachis rugosa, \\ 
Columbellidae Columbella haemastoma, Columbellidae Columbella labiosa, \\ 
Columbellidae Columbella major, Columbellidae Columbella paytensis, \\ 
Columbellidae Columbella socorroensis, Columbellidae Columbella strombiformis, \\ 
Columbellidae euplica varians, Columbellidae mitrella elegans baiyeli, \\ 
Columbellidae mitrella guttata, Columbellidae pyrene ocellata, \\ 
Conidae Conus brunneus, Conidae Conus dalli, Conidae Conus fergusoni, \\ 
Conidae Conus princeps, Conidae Conus regularis, \\ 
Conidae conus chaldaeus, Conidae conus ebraeus, Conidae conus gladiator, \\ 
Conidae conus nux, Conidae conus purpurascens, \\ 
Conidae conus scalaris, Conidae conus tessulatus, \\ 
Cymatiidae Monoplex gemmatus, Cymatiidae Monoplex pilearis, \\ 
Cymatiidae Monoplex vestitus, Cymatiidae Monoplex wiegmanni, \\ 
Cypraeidae Macrocypraea cervinetta, Cypraeidae Pseudozonaria arabicula, \\ 
Cypraeidae Pseudozonaria robertsi, Cypraeidae cypraea cervinetta, \\ 
Cypraeidae cypraea robertsi, Fasciolariidae Granolaria salmo, \\ 
Fasciolariidae Leucozonia cerata, Fasciolariidae Triplofusus princeps, \\ 
Fasciolariidae leucozonia rudis, Fasciolariidae opeatostoma pseudodon, \\ 
Ficidae Ficus ventricosa, Fissurellidae fissurella virescens, \\ 
Hipponicidae Antisabia panamensis, Hipponicidae Cheilea corrugata, \\ 
Janthinidae janthina janthina, Littorinidae Echinolittorina aspera, \\ 
Lottiidae Lottia fascicularis, Lottiidae Lottia filosa, \\ 
Lottiidae Lottia mesoleuca, Melongenidae Melongena patula, \\ 
Mitridae Neotiara lens, Mitridae Strigatella tristis, \\ 
Modulidae Trochomodulus catenulatus, Muricidae Hexaplex brassica, \\ 
Muricidae Hexaplex princeps, Muricidae Muricanthus radix, \\ 
Muricidae Neorapana muricata, Muricidae Phyllonotus regius, \\ 
Muricidae Plicopurpura collumelaris, Muricidae Plicopurpura columelaris, \\ 
Muricidae Stramonita haemastoma, Muricidae Thaisella kiosquiformis, \\ 
Muricidae Zetecopsis zeteki, Muricidae acanthais brevidentata, \\ 
Muricidae cymia tectum, Muricidae stramonita biserialis, \\ 
Muricidae thais tuberosa, Muricidae vasula melones, \\ 
Nassariidae nassarius erythraeus, Nassariidae nassarius gemmuliferus, \\ 
Naticidae Polinices uber, Naticidae mammilla simiae, \\ 
Naticidae natica fasciata, Naticidae notocochlis chemnitzii, \\ 
Neritidae Nerita funiculata, Neritidae nerita scabricosta, \\ 
Olividae Agaronia nica, Olividae Agaronia propatula, \\ 
Olividae Agaronia testacea, Olividae Oliva incrassata, \\ 
Olividae Oliva polpasta, Olividae Oliva porphyria, \\ 
Olividae Olivella volutella, Olividae Pachyoliva semistriata, \\ 
Ovulidae Jenneria pustulata, Personidae Distorsio decussata, \\ 
Pisaniidae Engina tabogaensis, Pisaniidae Gemophos ringens, \\ 
Pisaniidae Hesperisternia vibex, Pisaniidae Solenosteira gatesi, \\ 
Planaxidae Supplanaxis planicostatus, Planaxidae planaxis obsoletus, \\ 
Siphonariidae Siphonaria gigas, Siphonariidae Siphonaria maura, \\ 
Strombidae Lobatus peruvianus, Strombidae Persististrombus granulatus, \\ 
Strombidae Strombus alatus, Strombidae Strombus gracilior, \\ 
Strombidae Titanostrombus galeatus, Strombidae strombus granulatus, \\ 
Tegulidae Tegula pellisserpentis, Tegulidae tegula panamensis, \\ 
Tonnidae Malea ringens, Triviidae trivia sanguinea, \\ 
Turbinellidae vasum caestus, Turbinidae Arene olivacea, \\ 
Turbinidae Turbo saxosus, Turbinidae Uvanilla buschii, \\ 
Turritellidae Caviturritella leucostoma \\ 
\bottomrule
\end{longtable}

\vspace{1cm}

\begin{longtable}{p{\textwidth}}
\caption{Seashell Species - Caribbean Bivalves} \label{tab:caribbean_bivalves} \\ 
\toprule
\textbf{Species} \\ 
\midrule
Arcidae Anadara brasiliana, Arcidae Anadara chemnitzii, Arcidae Anadara notabilis, \\ 
Arcidae Anadara transversa, Arcidae Arca imbricata, Arcidae Arca zebra, \\ 
Arcidae Barbatia cancellaria, Arcidae Barbatia candida, Arcidae Barbatia dominguensis, \\ 
Arcidae Fugleria tenera, Arcidae Lamarcka imbricata, Cardiidae Acrosterigma magnum, \\ 
Cardiidae Dallocardia muricata, Cardiidae Laevicardium pictum, Cardiidae Papyridea semisulcata, \\ 
Cardiidae Papyridea soleniformis, Cardiidae Trachycardium isocardia, \\ 
Cardiidae Trachycardium magnum, Cardiidae Trachycardium muricatum, \\ 
Carditidae Carditamera gracilis, Chamidae Arcinella arcinella, Chamidae Chama congregata, \\ 
Chamidae Chama florida, Chamidae Chama macerophylla, Chamidae Chama sinuosa, \\ 
Chamidae Pseudochama cristella, Chamidae Pseudochama radians, Corbulidae Corbula caribaea, \\ 
Corbulidae Corbula contracta, Corbulidae Juliacorbula aquivalvis, Cyrenidae Polymesoda arctata, \\ 
Donacidae Donax denticulatus, Donacidae Donax striatus, Dreissenidae Mytilopsis sallei, \\ 
Glycymerididae Glycymeris undata, Glycymerididae Tucetona pectinata, \\ 
Isognomonidae Isognomon alatus, Isognomonidae Isognomon bicolor, Isognomonidae Isognomon radiatus, \\ 
Limidae Ctenoides scaber, Limidae Lima caribaea, Limidae Lima caribea, \\ 
Limidae Lima lima, Limidae Limaria pellucida, Lucinidae Anodontia alba, \\ 
Lucinidae Callucina keenae, Lucinidae Clathrolucina costata, Lucinidae Codakia orbicularis, \\ 
Lucinidae Ctena orbiculata, Lucinidae Divalinga quadrisulcata, \\ 
Lucinidae Divaricella quadrisulcata, Lucinidae Lucinisca centrifuga, \\ 
Lucinidae Phacoides pectinatus, Mactridae Mactrellona alata, Mactridae Mactroma fragilis, \\ 
Mactridae Mactrotoma fragilis, Mactridae Mulinia cleryana, Margaritidae Pinctada imbricata, \\ 
Mytilidae Botula fusca, Mytilidae Brachidontes exustus, Mytilidae Lioberus castanea, \\ 
Mytilidae Modiolus americanus, Nuculanidae Adrana lancea, \\ 
Nuculanidae Adrana tellinoides, Ostreidae Crassostrea rhizophorae, \\ 
Ostreidae Crassostrea virginica, Ostreidae Dendostrea frons, Pectinidae Spathochlamys benedicti, \\ 
Pectinidae Antillipecten antillarum, Pectinidae Argopecten gibbus, \\ 
Pectinidae Argopecten irradians, Pectinidae Argopecten irradians amplicostatus, \\ 
Pectinidae Caribachlamys ornata, Pectinidae Caribachlamys sentis, \\ 
Pectinidae Chlamys ornata, Pectinidae Euvola laurentii, Petricolidae Petricola bicolor, \\ 
Pinnidae Atrina seminuda, Pinnidae Pinna carnea, Plicatulidae Plicatula gibbosa, \\ 
Psammobiidae Asaphis deflorata, Psammobiidae Psammotella cruenta, \\ 
Pteriidae Pteria colymbus, Semelidae Semele proficua, Semelidae Semele purpurascens, \\ 
Solecurtidae Solecurtus cumingianus, Solecurtidae Tagelus divisus, \\ 
Spondylidae Spondylus butleri, Tellinidae Arcopagia fausta, Tellinidae Eurytellina angulosa, \\ 
Tellinidae Eurytellina nitens, Tellinidae Eurytellina punicea, Tellinidae Johnsonella fausta, \\ 
Tellinidae Laciolina laevigata, Tellinidae Merisca cristallina, Tellinidae Psammotreta brevifrons, \\ 
Tellinidae Scissula similis, Tellinidae Serratina aequistriata, Tellinidae Strigilla carnaria, \\ 
Tellinidae Strigilla dichotoma, Tellinidae Strigilla pisiformis, \\ 
Tellinidae Strigilla pseudocarnaria, Tellinidae Tellina punicea, \\ 
Tellinidae Tellina radiata, Tellinidae Tellinella listeri, Tellinidae Tellinella listeri, \\ 
Ungulinidae Diplodonta punctata, Ungulinidae Phlyctiderma semiasperum, \\ 
Veneridae Anomalocardia brasiliana, Veneridae Anomalocardia flexuosa, \\ 
Veneridae Chione cancellata, Veneridae Chione intapurpurea, \\ 
Veneridae Chione paphia, Veneridae Chionopsis intapurpurea, \\ 
Veneridae Dosinia concentrica, Veneridae Globivenus rigida, \\ 
Veneridae Gouldia cerina, Veneridae Hysteroconcha circinata, \\ 
Veneridae Hysteroconcha dione, Veneridae Lamelliconcha circinatus, \\ 
Veneridae Lirophora paphia, Veneridae Macrocallista maculata, \\ 
Veneridae Megapitaria maculata, Veneridae Pitar albidus, Veneridae Pitar fulminatus, \\ 
Veneridae Tivela mactroides, Veneridae Transennella cubaniana, \\ 
Veneridae Transennella stimpsoni, Veneridae Ventricola rigida \\ 
\bottomrule
\end{longtable}

\begin{longtable}{p{\textwidth}}
\caption{Seashell Species - Pacific Bivalves} \label{tab:pacific_bivalves} \\ 
\toprule
\textbf{Species} \\ 
\midrule
Arcidae Acar gradata, Arcidae Acar rostae, Arcidae Anadara similis, Arcidae Anadara tuberculosa, \\ 
Arcidae Arca pacifica, Arcidae Barbatia lurida, Arcidae Barbatia reeveana, \\ 
Arcidae Lamarcka mutabilis, Arcidae Larkinia grandis, Arcidae Larkinia multicostata, \\ 
Cardiidae Acrosterigma pristipleura, Cardiidae Americardia biangulata, Cardiidae Americardia planicostata, \\ 
Cardiidae Dallocardia senticosum, Cardiidae Laevicardium substriatum, Cardiidae Papyridea aspersa, \\ 
Cardiidae Trachycardium procerum, Carditidae Cardita crassicosta, Carditidae Carditamera affinis, \\ 
Carditidae Carditamera radiata, Carditidae Cardites crassicostatus, Carditidae Cardites laticostatus, \\ 
Carditidae Strophocardia megastropha, Chamidae Chama buddiana, Chamidae Chama coralloides, \\ 
Chamidae Chama echinata, Corbulidae Caryocorbula amethystina, Corbulidae Caryocorbula biradiata, \\ 
Corbulidae Caryocorbula nasuta, Corbulidae Caryocorbula ovulata, Crassatellidae Eucrassatella gibbosa, \\ 
Cyrenoididae Polymesoda inflata, Donacidae Donax carinatus, Donacidae Donax dentifer, \\ 
Donacidae Iphigenia altior, Glycymerididae Axinactis delessertii, Glycymerididae Axinactis inaequalis, \\ 
Glycymerididae Tucetona multicostata, Gryphaeidae Hyotissa hyotis, Isognomonidae Isognomon recognitus, \\ 
Limidae Limaria tetrica, Lucinidae Codakia distinguenda, Lucinidae Ctena galapagana, \\ 
Lucinidae Ctena mexicana, Lucinidae Divalinga eburnea, Mactridae Harvella elegans, \\ 
Mactridae Mactrellona clisia, Mactridae Mactrellona exoleta, Mactridae Mactrellona subalata, \\ 
Mactridae Mulinia pallida, Mytilidae Brachidontes puntarenensis, Mytilidae Leiosolenus aristatus, \\ 
Mytilidae Leiosolenus plumula, Mytilidae Modiolus capax, Mytilidae Mytella guyanensis, \\ 
Noetiidae Noetia reversa, Ostreidae Crassostrea columbiensis, Ostreidae Crassostrea corteziensis, \\ 
Ostreidae Crassostrea gigas, Ostreidae Saccostrea palmula, Ostreidae Striostrea prismatica, \\ 
Pectinidae Argopecten ventricosus, Pectinidae Nodipecten subnodosus, Pinnidae Atrina maura, \\ 
Pinnidae Pinna rugosa, Psammobiidae Gari helenae, Psammobiidae Heterodonax pacificus, \\ 
Psammobiidae Sanguinolaria tellinoides, Pteriidae Pinctada mazatlanica, Pteriidae Pteria sterna, \\ 
Semelidae Semele bicolor, Semelidae Semele elliptica, Semelidae Semele formosa, \\ 
Semelidae Semele purpurascens, Semelidae Semele verrucosa, Solecurtidae Tagelus affinis, \\ 
Solecurtidae Tagelus peruanus, Solecurtidae Tagelus peruvianus, Spondylidae Spondylus limbatus, \\ 
Tellinidae Eurytellina regia, Tellinidae Iridona subtrigona, Tellinidae Psammotreta pura, \\ 
Tellinidae Strigilla chroma, Tellinidae Strigilla dichotoma, Tellinidae Strigilla disjuncta, \\ 
Tellinidae Strigilla serrata, Ungulinidae Zemysina subquadrata, Veneridae Chione subimbricata, \\ 
Veneridae Cyclinella producta, Veneridae Cyclinella subquadrata, Veneridae Dosinia dunkeri, \\ 
Veneridae Dosinia ponderosa, Veneridae Hysteroconcha lupanaria, Veneridae Hysteroconcha multispinosus, \\ 
Veneridae Hysteroconcha roseus, Veneridae Iliochione subrugosa, Veneridae Lamelliconcha tortuosus, \\ 
Veneridae Lamelliconcha unicolor, Veneridae Leukoma asperrima, Veneridae Leukoma ecuadoriana, \\ 
Veneridae Leukoma grata, Veneridae Leukoma histrionica, Veneridae Megapitaria aurantiaca, \\ 
Veneridae Megapitaria squalida, Veneridae Periglypta multicostata, Veneridae Tivela byronensis, \\ 
Veneridae Tivela planulata \\ 
\bottomrule
\end{longtable}

{
    \small
    \clearpage
    \bibliographystyle{ieeenat_fullname}
    \bibliography{main}
}

\end{document}